# Unsupervised Domain Adaptation for 3D LiDAR Semantic Segmentation Using Contrastive Learning and Multi-Model Pseudo Labeling

**ABHISHEK KAUSHIK[1], NORBERT HAALA[1] & UWE SÖRGEL[1]**

*Addressing performance degradation in 3D LiDAR semantic segmentation due to domain shifts (e.g., sensor type, geographical location) is crucial for autonomous systems, yet manual annotation of target data is prohibitive. This study addresses the challenge using Unsupervised Domain Adaptation (UDA) and introduces a novel two-stage framework to tackle it. Initially, unsupervised contrastive learning at the segment level is used to pre-train a backbone network, enabling it to learn robust, domain-invariant features without labels. Subsequently, a multi-model pseudo-labeling strategy is introduced, utilizing an ensemble of diverse state-of-the-art architectures (including projection, voxel, hybrid, and cylinder-based methods). Predictions from these models are aggregated via hard voting to generate high-quality, refined pseudo-labels for the unlabeled target domain, mitigating single-model biases. The contrastively pre-trained network is then fine-tuned using these robust pseudo-labels. Experiments adapting from SemanticKITTI to unlabeled target datasets (SemanticPOSS, SemanticSlamantic) demonstrate significant improvements in segmentation accuracy compared to direct transfer and single-model UDA approaches. These results highlight the effectiveness of combining contrastive pre-training with refined ensemble pseudo-labeling for bridging complex domain gaps without requiring target domain annotations.*

## 1 Introduction

The ability to semantically understand 3D environments captured by LiDAR sensors is paramount for the advancement of autonomous systems, including self-driving vehicles, robotic navigation, and smart infrastructure. By assigning semantic labels to individual 3D points, these systems can effectively perceive and interact with their surroundings. Deep learning models have demonstrated remarkable success in this domain, achieving high segmentation accuracy when trained on large-scale, richly annotated datasets: SemanticKITTI (BEHLEY et al. 2019; Geiger et al. 2012), SemanticPOSS (PAN et al. 2020). However, the manual annotation of 3D point clouds is a laborious and expensive undertaking, posing a significant bottleneck for the widespread deployment and scalability of these sophisticated models in diverse real-world applications.

A significant challenge hindering widespread adoption is the domain gap, which causes substantial performance drops when models trained on a specific labeled source dataset are applied to unlabeled target data with differing characteristics. These domain shifts stem from various factors. Sensor variations, such as differences in the number of LiDAR beams, resulting point density, and scanning patterns, significantly alter the raw data structure. Furthermore, sensor viewpoint differences, arising from varied mounting configurations – for example, sensors mounted on diverse moving platforms like bikes, cars, and trucks (as seen in dataset like SemanticSlamantic) Environmental conditions (e.g., snow vs. sunny) and geographical location also play crucial roles;

---

[1] Universität Stuttgart, Institut für Photogrammetrie und Geoinformatik, Geschwister-Scholl-Straße 24D, D-70174 Stuttgart, E-Mail: [abhishek.kaushik, norbert.haala, uwe.soergel]@ifp.uni-stuttgart.de

for instance, a model trained solely on SemanticKITTI (BEHLEY et al. 2019 ; Geiger et al. 2012), collected in Germany, will inherently struggle when applied to SemanticPOSS (PAN et al. 2020), captured in China, due to differences in urban layouts, object frequencies, and architectural styles. The combination of these factors, particularly the shift from data acquisition with varying sensor types and geographical contexts, necessitates effective domain adaptation techniques.

Unsupervised Domain Adaptation (UDA) offers a promising pathway to overcome the limitations imposed by the need for extensive manual annotations in the target domain. By enabling the transfer of knowledge learned from a labeled source domain to an unlabeled target domain, UDA holds the key to building more robust and generalizable 3D perception systems. Among the various UDA strategies, contrastive learning has emerged as a powerful approach for learning domain-invariant feature representations by encouraging similar data points to cluster together in the feature space while separating dissimilar ones (NUNES et al. 2022). Another prevalent technique is pseudo-labeling, where predictions from a source-trained model are used as surrogate labels for the unlabeled target data, facilitating further training and adaptation.

In this study, we address the challenging problem of unsupervised domain adaptation for 3D LiDAR semantic segmentation, specifically focusing on bridging the domain gap arising from sensor variations (number of beams, point density, viewpoint/platform dynamics) and geographical location differences. We tackle the scenario of adapting from the well-annotated, primarily SemanticKITTI (BEHLEY et al. 2019; Geiger et al. 2012) dataset to unlabeled target data potentially captured from different sensors, moving platforms (e.g., bikes, cars, trucks), and distinct geographical locations (like those represented in SemanticPOSS (PAN et al. 2020)). To effectively bridge this complex domain gap, we propose a novel UDA framework that synergistically combines contrastive learning for initial domain-invariant feature learning with a robust multi-model pseudo-labeling strategy for refinement.

Recognizing that different state-of-the-art 3D semantic segmentation architectures possess unique strengths and weaknesses, we first study the effect of diverse architectures – encompassing projection-based (e.g., RangeNet++ (MILIOTO et al. 2019)), partition-based (e.g., Cylinder3D (ZHOU et al. 2021), MinkUNet (CHOY et al. 2019)), and point-voxel hybrid methods (e.g., SPVNAS (TANG et al. 2020)) – on the unseen target data, analyzing their class-wise performance. Based on this analysis, our approach leverages the complementary strengths of these multiple architectures to generate and refine pseudo-labels for the target domain. By employing an ensemble-based voting mechanism, we aggregate predictions to mitigate the inherent noise and biases of single-model pseudo-labels, thereby improving label quality. Our training methodology involves pre-training the model using contrastive learning (NUNES et al. 2022) to align features across domains, followed by fine-tuning using the refined pseudo-labels generated by the multi-model ensemble. This work also aims to explicitly demonstrate the critical importance of high-quality pseudo-labels for achieving strong final segmentation performance.

This work presents a novel UDA framework for 3D LiDAR semantic segmentation, combining contrastive learning pre-training for domain-invariant features with a multi-model pseudo-labeling approach for fine-tuning. We address domain shifts due to sensor variations and geographical differences by adapting from single source data (SemanticKITTI (BEHLEY et al. 2019; Geiger et al. 2012)) to unlabeled target datasets (SemanticPOSS (PAN et al. 2020) & SemanticSlamantic). Experimental results show significant improvements in segmentation accuracy compared to direct transfer and single-model UDA baselines, highlighting the importance of refined pseudo-labels in the adaptation process.

## 2 Theory & Related Works

This section provides the theoretical background and reviews relevant prior work that forms the foundation of our research. We first examine the domain of 3D semantic segmentation, highlighting the significant challenge posed by domain shift and the various Unsupervised Domain Adaptation (UDA) techniques developed to address this issue. Subsequently, we delve deeper into the principles and methodologies of Domain Adaptation, elaborating on the strategies employed to mitigate the performance degradation of models when applied to unseen target domains.

### 2.1 3D Semantic Segmentation

3D semantic segmentation is a fundamental task in computer vision, particularly for autonomous systems, involving the per-point classification of LiDAR-generated point clouds to understand 3D environments. This process faces challenges due to the sparse, irregular, and unstructured nature of point cloud data. Various approaches address these challenges: Point-based methods directly process raw points, preserving fine-grained detail, with seminal works including PointNet (Qi et al. 2017), which introduced permutation-invariant processing, PointNet++ (Qi et al. 2017), which added hierarchical feature learning, KPConv (Thomas et al. 2019), which proposed point convolutions with Euclidean-space kernels, and RandLA-Net (Hu et al. 2020), which efficiently handles large-scale clouds via random sampling and local feature aggregation. Projection-based methods transform 3D data into 2D representations (e.g., range images) to leverage 2D CNNs, exemplified by SqueezeSeg (Wu et al. 2018), and its variants improving robustness, RangeNet++ (Milioto et al. 2019) achieving real-time performance with efficient backbones and post-processing, and LENet (Ding et al. 2023) employing multi-scale attention and efficient upsampling. Partition-based methods divide the 3D space into manageable regions like voxels or cylinders; examples include Cylinder3D (Zhou et al. 2021), which uses cylindrical partitioning and convolutions tailored for driving scenes, and MinkUNet (Choy et al. 2019), which employs efficient sparse convolutions on voxelized representations using the Minkowski Engine. Hybrid methods combine voxel- and point-based representations to leverage both structural context and fine-grained detail. SPVNAS (TANG et al. 2020), for instance, integrates sparse 3D convolutions with point-wise refinement to achieve a balance between accuracy and efficiency. Such approaches exploit the strengths of both domains, enhancing segmentation performance on complex scenes.

### 2.2 Domain Adaptation

The performance of 3D semantic segmentation models can significantly degrade due to domain shift, where discrepancies between the training (source) and deployment (target) data distributions arise from factors like differing sensor types, environmental conditions (e.g., urban vs. rural, weather), or object characteristics. Unsupervised Domain Adaptation (UDA) seeks to mitigate this issue by adapting models trained on labeled source data (often synthetic) to unlabeled target data (often real-world) without requiring costly target domain annotations. Various UDA strategies have been developed. Some approaches leverage pseudo-labeling, often within teacher-student frameworks, as seen in methods like ST3D (YANG et al. 2021), and ST3D++ (YANG et al. 2021), primarily for object detection. For semantic segmentation, techniques include cross-modal learning like xMUDA (JARITZ et al. 2020), which enforces consistency between 2D image and 3D point cloud predictions. Input-level mixing strategies, such as CoSMix (SALTORI et al. 2022), use compositional operations based on semantics within a dual-branch architecture to blend source and target point clouds. Contrastive learning has also been applied; SegContrast (NUNES et al. 2022) learns structural representations by contrasting class-agnostic point cloud segments, while PointContrast (XIE et al. 2020), performs dense contrastive learning between points in augmented views of the same cloud, particularly demonstrated for indoor scenes. The overarching goal of these domain adaptation techniques is to enhance model generalization and robustness across diverse operational domains for reliable real-world deployment.

## 3  Methodology

This research tackles the challenge of unsupervised domain adaptation (UDA) for 3D LiDAR semantic segmentation, focusing on bridging the domain gap caused by differences in sensor specifications and environmental conditions between a labeled source domain and an unlabeled target domain. The proposed methodology adopts a two-stage approach: first, an unsupervised contrastive learning framework is used to pre-train a 3D backbone network on the source & target domain, enabling it to learn robust and domain-invariant features. Second, a supervised fine-tuning stage is performed using high-quality pseudo-labels generated by an ensemble of diverse 3D semantic segmentation architectures, leading to accurate semantic predictions on the target domain.

**Unsupervised Pre-training via Contrastive Learning**
The first stage focuses on learning discriminative and domain-invariant feature representations from the unlabeled target domain data in a self-supervised manner. Inspired by SegContrast (NUNES et al. 2022) this stage relies on contrastive learning at the segment level. The input point cloud is first passed through a 3D backbone network—typically a sparse convolutional neural network—to extract point-wise features that encode local geometric and semantic information.

To enable contrastive learning at the segment level, the point cloud is first segmented into distinct structural components. This segmentation process begins with ground plane removal using the RANSAC (Random Sample Consensus) algorithm (FISCHLER et al. 1981), which robustly fits a plane model and removes ground points. Eliminating the ground plane helps the model focus on more meaningful structures in the scene.

The remaining non-ground points are then clustered using DBSCAN (Density-Based Spatial Clustering of Applications with Noise) (ESTER et al. 1996), a density-based clustering algorithm that groups closely packed points into segments while identifying sparse regions as outliers. To prevent memory overflow during training and avoid over-segmentation, only the top $\delta$ clusters (ranked by number of points) are retained, and a minimum point threshold $\epsilon$ is enforced for valid clusters. This results in a set of meaningful structural segments from the input point cloud.

To train the model with contrastive loss, two augmented versions of the same point cloud are generated using a diverse set of augmentations. These include random cropping (cuboid extraction), rotation, scaling, flipping, cuboid dropout, point jittering, and fine-grained rotation perturbations. Each augmented view is passed through the shared backbone network to extract features. For each segment, the corresponding point-wise features are aggregated using a projection head followed by dropout and global max-pooling, resulting in a compact segment-level feature vector. This vector is further transformed by another projection head to obtain the final representation used in contrastive learning. The InfoNCE loss (VAN DEN OORD et al. 2019) is then applied, encouraging the model to assign high similarity to positive pairs (same segment across different views) and low similarity to negative pairs (different segments). Through this process, the model learns segment-level features that are invariant to transformations and effective for semantic discrimination, all without relying on manual annotations.

**Pseudo-Label Generation with Ensemble Voting**

The second stage focuses on generating pseudo-labels for the unlabeled target domain using an ensemble of pre-trained 3D semantic segmentation models. The ensemble includes diverse state-of-the-art architectures, spanning projection-based methods (e.g., LENet (DING et al. 2023)), voxel-based approaches (e.g., MinkUNet (Choy et al. 2019)), hybrid point+voxel techniques (e.g., SPVNAS (TANG et al. 2020)), and cylinder-based models (e.g., Cylinder3D (ZHOU et al. 2021)). Each model provides a full semantic prediction on the target domain data based on its respective strengths and learned features.

To obtain a unified and robust pseudo-label for each point, we implement a hard voting strategy. For every point in the point cloud, the predicted semantic labels from all ensemble models are collected. The final pseudo-label is assigned based on the majority vote, mitigating individual model biases and errors. This ensemble voting mechanism yields a set of pseudo-labels that is generally more consistent and accurate than predictions from any single model.

A designated subset of the unlabeled target data is reserved for this ensemble-based pseudo-label generation step, ensuring that the fine-tuning process benefits from the most reliable pseudo-annotations available.

**Supervised Fine-tuning with Pseudo-Labels**

In the final stage, the pre-trained backbone network is fine-tuned on the unlabeled target data using the pseudo-labels as surrogate ground truth. This step treats the high-quality pseudo-labels as supervision targets and trains the model using a standard supervised objective, such as cross-entropy loss. The backbone, having already learned generalizable structural features during contrastive pre-training, is now guided to associate these features with specific semantic categories present in the target domain.

This fine-tuning allows the network to refine its understanding of the target domain, effectively adapting to its unique semantic distributions. The result is a model that leverages both unsupervised structural understanding and supervised semantic alignment, leading to improved semantic segmentation performance on the unlabeled target data.

## 4  Experiments & Results

This section details the experiments conducted to validate our Unsupervised Domain Adaptation (UDA) framework. We utilized SemanticKITTI as the source dataset and SemanticSlamantic (unlabeled) and SemanticPOSS (labeled) as target datasets to evaluate performance across sensor and geographical domain shifts. The results assess the effectiveness of our contrastive pre-training and ensemble pseudo-labeling approach in bridging these gaps without target annotations.

## 4.1 Datasets

Our study focuses on 3d semantic segmentation, utilizing SemanticKITTI (BEHLEY et al. 2019) as the source dataset and SemanticSlamantic as the unlabeled target dataset. Both datasets consist of outdoor scenes collected in Germany. SemanticKITTI, derived from the KITTI Vision Benchmark (GEIGER et al. 2012), provides annotated 3D scans acquired with a 64-beam Velodyne LiDAR mounted on a car, featuring a vertical field of view (FOV) of [−2◦, −24.8◦] (see Tab. 1). SemanticSlamantic comprises a larger set of unannotated scans captured using a 64-beam Ouster LiDAR mounted on various vehicles (bike, car, truck), with a wider vertical FOV of [−22.5◦, +22.5◦] (see Tab. 1). The domain gap between these datasets is primarily due to variations in sensor mounting positions and sensor types. To quantitatively evaluate our framework's performance, we employ SemanticPOSS (PAN et al. 2020), which offers annotated scans from a Jeep-mounted 40-beam Pandora LiDAR in China (see Tab. 1). The domain shift between SemanticKITTI and SemanticPOSS is attributed to differences in sensor types and geographical location. Addressing these domain gaps is central to our methodology for achieving robust semantic segmentation across diverse LiDAR data.

Tab. 1: Summary of the datasets employed.

| Datasets | Annotated | #Scans | Lasers | Mounted On | Location | Sensor |
|---|---|---|---|---|---|---|
| **SemanticKITTI** (BEHLEY et al. 2019) | Yes | 23,201 | 64-beam | Car | Germany | Velodyne |
| **SemanticSlamantic** | No | 68,951 | 64-beam | Bike, Car, Truck | Germany | Ouster |
| **SemanticPOSS** (PAN et al. 2020) | Yes | 2,988 | 40-beam | Jeep | China | Pandora |

## 4.2 On SemanticSlamantic

In our experiments aimed at identifying an optimal architecture for our target datasets (SemanticSlamantic & SemanticPOSS (PAN et al. 2020)), we conducted supervised learning on the SemanticKITTI (BEHLEY et al. 2019) dataset and evaluated the performance on its validation split. As indicated in the results presented in Table 1, no single architecture demonstrated superior performance across all semantic classes. To address this, we explored enhancements for specific architectures; for instance, applying kNN (MILIOTO et al. 2019) post-processing to the LeNet architecture led to improved performance, particularly for certain classes. Additionally, we investigated the impact of polar mix augmentation (LPX) (XIAO et al. 2022) across various models, and the results in Table 2 consistently show that incorporating LPX yielded gains in both overall accuracy and average IoU, as well as for many individual class categories.

Tab. 2: Semantic segmentation performance of different 3D point cloud architectures on the SemanticKITTI validation set, evaluating the impact of kNN post-processing on LeNet and the application of polar mix augmentation (LPX) across various copies. Performance is reported in terms of overall accuracy (Acc avg) and mean Intersection-over-Union (IoU avg), along with the IoU for each semantic class

| Method | Acc avg | IoU avg | car | bicycle | motorcycle | truck | other-vehicle | person | bicyclist | road | parking | sidewalk | building | fence | vegetation | trunk | terrain | pole | traffic-sign |
|---|---|---|---|---|---|---|---|---|---|---|---|---|---|---|---|---|---|---|---|
| LeNet | 0.89 | 0.58 | 0.87 | 0.40 | 0.53 | 0.79 | 0.52 | 0.66 | 0.71 | 0.95 | 0.41 | 0.82 | 0.81 | 0.50 | 0.84 | 0.60 | 0.65 | 0.51 | 0.45 |
| LeNet+kNN | 0.88 | 0.62 | 0.92 | 0.45 | 0.57 | 0.81 | 0.54 | 0.74 | 0.82 | 0.95 | 0.43 | 0.82 | 0.83 | 0.53 | 0.79 | 0.67 | 0.65 | 0.64 | 0.47 |
| Cylinder3D | 0.92 | 0.63 | 0.97 | 0.42 | 0.66 | 0.88 | 0.56 | 0.72 | 0.89 | 0.94 | 0.49 | 0.82 | 0.91 | 0.49 | 0.87 | 0.57 | 0.72 | 0.65 | 0.48 |
| Cylinder3D+LPX | 0.92 | 0.68 | 0.97 | 0.59 | 0.78 | 0.87 | 0.75 | 0.82 | 0.91 | 0.94 | 0.55 | 0.81 | 0.92 | 0.61 | 0.88 | 0.70 | 0.74 | 0.62 | 0.51 |
| MinkNet | 0.92 | 0.63 | 0.97 | 0.19 | 0.63 | 0.86 | 0.64 | 0.68 | 0.84 | 0.94 | 0.49 | 0.83 | 0.92 | 0.63 | 0.87 | 0.66 | 0.75 | 0.64 | 0.49 |
| MinkNet+LPX | 0.92 | 0.69 | 0.98 | 0.54 | 0.82 | 0.86 | 0.79 | 0.78 | 0.89 | 0.94 | 0.50 | 0.81 | 0.92 | 0.65 | 0.89 | 0.68 | 0.72 | 0.66 | 0.53 |
| SPVNAS | 0.93 | 0.64 | 0.97 | 0.30 | 0.63 | 0.84 | 0.67 | 0.71 | 0.86 | 0.95 | 0.49 | 0.83 | 0.92 | 0.67 | 0.88 | 0.65 | 0.77 | 0.64 | 0.49 |
| SPVNAS+LPX | 0.93 | 0.69 | 0.97 | 0.56 | 0.81 | 0.81 | 0.77 | 0.81 | 0.90 | 0.95 | 0.50 | 0.83 | 0.92 | 0.67 | 0.88 | 0.70 | 0.74 | 0.66 | 0.53 |

The performance of the proposed UDA framework significantly benefits from the ensemble voting strategy used for pseudo-label generation. Initially, diverse state-of-the-art models given in Table 2, each trained on the labeled SemanticKITTI source dataset, were used to generate initial pseudo-labels for the unlabeled target domain. While fine-tuning with pseudo-labels from individual architectures showed improvements over direct transfer, the final ensemble voting strategy aggregates predictions from these multiple models. This refinement process produces higher-quality, more robust pseudo-labels by mitigating the biases and errors inherent in any single architecture. Fine-tuning the contrastively pre-trained model using these refined ensemble pseudo-labels yielded the best overall performance. Notably, this approach proved particularly effective in segmenting challenging small object classes, such as 'person' and 'traffic signs', which were often misclassified or missed when using pseudo-labels from single models. Qualitative visualizations (as shown in Figure 1) clearly demonstrate the superior segmentation accuracy achieved with the refined ensemble pseudo-labels, particularly when contrasted with results derived from individual model pseudo-labels like those from Cylinder3D+LPX. Figure 1 specifically highlights how the ensemble approach leads to significant improvements for challenging classes, such as achieving highly accurate segmentation for 'person'. Crucially, these strong comparative results, showcasing the effectiveness of the ensemble strategy, were obtained entirely without utilizing any ground truth labeled point clouds from the target dataset. This confirms that the framework successfully bridges the domain gap caused by differing sensor configurations and mounting platforms between the source and target domains.

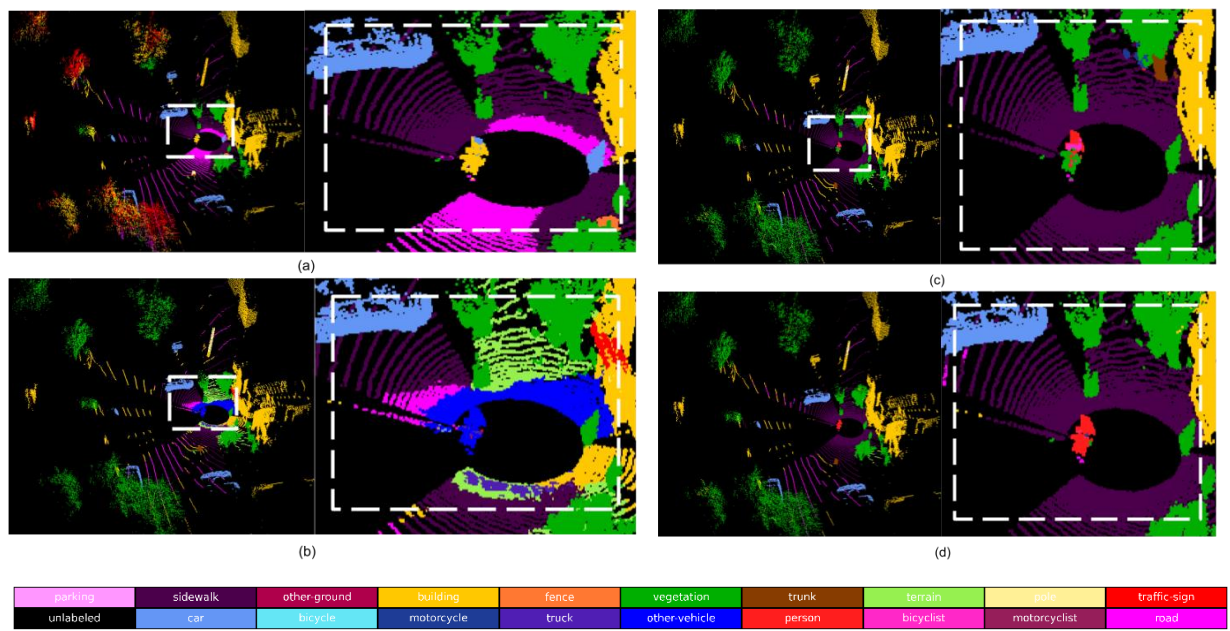

Fig.: 1 Qualitative comparison of 3D semantic segmentation results on the unlabeled target domain using the proposed Unsupervised Domain Adaptation framework. The panels display predictions obtained after fine-tuning the model with pseudo-labels generated from different sources: (a) Cylinder3D+LPX, (b) SPVNAS+LPX, (c) MinkUNet+LPX, and (d) the refined Ensemble voting strategy. Zoomed-in views are included to highlight performance differences on specific object classes like 'person', illustrating the superior accuracy of the ensemble approach (d).

## 4.3 On SemanticPOSS (PAN et al. 2020)

The above experiments underscore the critical importance of generating high-quality, refined pseudo-labels, with the ensemble voting strategy proving highly effective for this purpose. The robustness of the pseudo-labels generated during the refinement stage stems from leveraging multiple diverse architectures within the ensemble. As observed in initial experiments on the SemanticKITTI source dataset, each architecture possesses unique strengths and weaknesses, with no single model achieving optimal performance across all semantic classes. Utilizing predictions from several such models helps mitigate individual biases and errors. Furthermore, it's not guaranteed that the model exhibiting the highest IoU on the source domain will necessarily perform best when transferred to an unseen target domain, making the ensemble approach based on combined target predictions particularly valuable. Since the primary target dataset (SemanticSlamantic) lacks ground truth annotations needed for direct quantitative assessment of pseudo-label quality, the annotated SemanticPOSS dataset was utilized specifically for quantitatively evaluating the quality of the pseudo-labels generated by our methods. This quantitative analysis further reinforces the finding that the final segmentation performance of the adapted model is directly proportional to the quality of the pseudo-labels used for fine-tuning.

Tab.:3 Pseudo-label quality comparison: Models (trained on SemanticKITTI) vs. Ensemble on SemanticPOSS using IoU and Accuracy. Bold indicates best performance.

| Model | IoU avg | Acc avg | Building | Car/Vehicle | Fence | Ground | Person | Plants | Pole | Traffic-Sign | Trunk |
|---|---|---|---|---|---|---|---|---|---|---|---|
| MinkUNet | 0.486 | 0.825 | 0.716 | 0.347 | 0.286 | **0.820** | **0.383** | 0.745 | 0.354 | **0.225** | 0.504 |
| MinkUNet+LPX | 0.466 | 0.818 | 0.706 | **0.513** | 0.307 | 0.787 | 0.184 | 0.733 | 0.351 | 0.181 | 0.427 |
| Cyl3d+LPX | 0.357 | 0.753 | 0.628 | 0.356 | 0.192 | 0.806 | 0.024 | 0.649 | 0.257 | 0.055 | 0.248 |
| Cyl3d | 0.360 | 0.763 | 0.670 | 0.253 | 0.290 | 0.758 | 0.009 | 0.654 | 0.274 | 0.183 | 0.147 |
| SPVNAS | 0.475 | 0.822 | 0.730 | 0.314 | 0.309 | 0.806 | 0.210 | 0.735 | 0.395 | 0.212 | **0.560** |
| SPVNAS+LPX | 0.474 | 0.821 | 0.725 | 0.441 | 0.301 | 0.795 | 0.224 | 0.732 | 0.349 | 0.172 | 0.528 |
| Ensemble(Ours) | **0.505** | **0.842** | **0.756** | 0.470 | **0.353** | 0.817 | 0.259 | **0.757** | **0.383** | 0.222 | 0.530 |

To quantitatively evaluate the effectiveness of different pseudo-label generation strategies in bridging the significant domain gap between SemanticKITTI (Germany, Velodyne sensor) and SemanticPOSS (China, Pandora sensor), we assessed the quality of predictions from various state-of-the-art architectures on the annotated SemanticPOSS dataset. These architectures (MinkUNet, Cylinder3D, SPVNAS, with and without LPX augmentation), initially trained on SemanticKITTI, represent potential pseudo-label generators. The performance metrics, detailed in Table 3, reveal considerable variation among the individual models when applied to the target domain, highlighting that no single architecture is universally optimal across the combined geographical and sensor domain shift. For instance, while MinkUNet excels in predicting 'Ground' (0.820 IoU) and 'Person' (0.383 IoU), MinkUNet+LPX performs best for 'Car/Vehicle' (0.513 IoU), and SPVNAS achieves the top score for 'Trunk' (0.560 IoU). Notably, our proposed Ensemble voting strategy achieves the highest overall mean IoU (0.505) and average accuracy (0.842), clearly outperforming the predictions from any individual model baseline as shown in the table. This quantitative improvement demonstrates the efficacy of the ensemble approach in creating refined, higher-quality pseudo-labels that better generalize across domains. This experiment strongly validates the importance of utilizing a mixture of diverse architectures, leveraging their complementary strengths and weaknesses, and employing the ensemble voting mechanism for refinement. The resulting enhanced pseudo-label quality directly contributes to reducing the domain gap. Therefore, these validated, high-quality pseudo-labels generated via the ensemble method are subsequently used to fine-tune the contrastively pre-trained network, forming a critical step in our unsupervised domain adaptation framework.

## 5 Conclusion & Future Work

This work successfully demonstrated a novel UDA framework for 3D LiDAR semantic segmentation, effectively mitigating domain shifts from sensor and location variations by combining contrastive pre-training with ensemble-based pseudo-labeling. Our multi-model ensemble voting strategy proved crucial, generating higher-quality pseudo-labels than single models by leveraging diverse architectural strengths, leading to significant performance gains on target domains without ground truth. The results confirm the critical link between pseudo-label quality and adaptation success.

Future work could explore more sophisticated ensemble techniques (e.g., weighted voting, meta-learning), investigate iterative pseudo-label refinement, and extend the framework to online/continual adaptation scenarios. Further research could also involve integrating cross-modal data, testing robustness against wider domain gaps (extreme weather), and conducting deeper theoretical analyses of the ensemble benefits and the interplay between contrastive learning and pseudo-label fine-tuning.

## 6 Literature

GEIGER, A., LENZ, P. & URTASUN, R., 2012: Are we ready for autonomous driving? The KITTI vision benchmark suite, 2012 IEEE Conference on Computer Vision and Pattern Recognition, 3354–3361. https://doi.org/10.1109/CVPR.2012.6248074.

BEHLEY, J., GARBADE, M., MILIOTO, A., QUENZEL, J., KRAUS, D., NORTHOFF, S., STACHNISS, C., 2019: SemanticKITTI: A Dataset for Semantic Scene Understanding of LiDAR Sequences, Proceedings of the IEEE/CVF international conference on computer vision, 9703–9711.

PAN, Y., GAO, B., MEI, J., GENG, S., LI, C., & ZHAO, H., 2020: SemanticPOSS: A Point Cloud Dataset with Large Quantity of Dynamic Instances, 2020 IEEE Intelligent Vehicles Symposium, 687-693.

MILIOTO, A., VIZZO, I., BEHLEY, J., & STACHNISS, C., 2019: RangeNet++: Fast and Accurate LiDAR Semantic Segmentation, 2019 IEEE/RSJ International Conference on Intelligent Robots and Systems (IROS), 4213–4220.

ZHOU, H., ZHU, X., SONG, X., MA, Y., WANG, Z., LI, H., & LIN, D., 2021: Cylinder3D: Cylindrical and Asymmetrical 3D Convolution Networks for LiDAR Segmentation, 2021 Proceedings of the IEEE/CVF Conference on Computer Vision and Pattern Recognition.

TANG, H., LIU, Z., ZHAO, S., LIN, Y., LIN, J., WANG, H., & HAN, S., 2020: Searching Efficient 3D Architectures with Sparse Point-Voxel Convolution, 2020 Proceedings of the European Conference on Computer Vision (ECCV).

CHOY, C., GWAK, J., & SAVARESE, S., 2019: 4D Spatio-Temporal ConvNets: Minkowski Convolutional Neural Networks, 2019 Proceedings of the IEEE/CVF Conference on Computer Vision and Pattern Recognition.

NUNES, L., MARCUZZI, R., CHEN, J., BEHLEY, J., & STACHNISS, C., 2022: SegContrast: 3D Point Cloud Feature Representation Learning Through Self-Supervised Segment Discrimination, 2022 IEEE Robotics and Automation Letters.

SALTORI, C., GALASSO, F., FIAMENI, G., SEBE, N., RICCI, E., & POIESI, F., 2022: CoSMix: Compositional Semantic Mix for Domain Adaptation in 3D LiDAR Segmentation, 2022 Proceedings of the European Conference on Computer Vision (ECCV).

XIE, S., GU, J., GUO, D., QI, C.R., GUIBAS, L., & LITANY, O. 2020: PointContrast: Unsupervised Pre-training for 3D Point Cloud Understanding, 2020 Proceedings of the European Conference on Computer Vision (ECCV).

YANG, J., SHI, S., WANG, Z., LI, H., & QI, X., 2021: ST3D: Self-Training for Unsupervised Domain Adaptation on 3D Object Detection, 2021 Proceedings of the IEEE/CVF Conference on Computer Vision and Pattern Recognition.

YANG, J., SHI, S., WANG, Z., LI, H., & QI, X., 2021: ST3D++: Self-Training for Unsupervised Domain Adaptation on 3D Object Detection, IEEE Transactions on Pattern Analysis and Machine Intelligence.

JARITZ, M., VU, T.H., CHARETTE, R.D., WIRBEL, E. & PEREZ, P. 2020: xMUDA: Cross-Modal Unsupervised Domain Adaptation for 3D Semantic Segmentation, 2020 Proceedings of the IEEE/CVF Conference on Computer Vision and Pattern Recognition.

QI, C.R., SU, H., MO, K. & GUIBAS, L.J. 2017: PointNet: Deep Learning on Point Sets for 3D Classification and Segmentation, 2017 Proceedings of the IEEE/CVF Conference on Computer Vision and Pattern Recognition.

QI, C.R., YI, LI., SU, H., & GUIBAS, L.J. 2017: PointNet++: Deep Hierarchical Feature Learning on Point Sets in a Metric Space, 2017 Advances in Neural Information Processing Systems 30 (NIPS 2017).

THOMAS, H., QI, C.R., DESCHAUD, J.E., MARCOTEGUI, B., GOULETTE, F. & GUIBAS, L.J. 2019: KPConv: Flexible and Deformable Convolution for Point Clouds, 2019 Proceedings of the IEEE/CVF Conference on Computer Vision and Pattern Recognition.

HU, Q., YANG, B., XIE, L., ROSA, S., GUO, Y., WANG, Z., TRIGONI, N. & MARKHAM, A. 2020: RandLA-Net: Efficient Semantic Segmentation of Large-Scale Point Clouds, 2020 Proceedings of the IEEE/CVF Conference on Computer Vision and Pattern Recognition.

WU, B., WAN, A., YUE, X. & KEUTZER, K. 2018: SqueezeSeg: Convolutional Neural Nets with Recurrent CRF for Real-Time Road-Object Segmentation from 3D LiDAR Point Cloud, 2018 IEEE International Conference on Robotics and Automation (ICRA).

DING, B. 2023: Lenet: Lightweight and efficient lidar semantic segmentation using multiscale convolution attention, 2023, https://arxiv.org/abs/2301.04275.

FISCHLER, M.A. & BOLLES, R.C. 1981: Random sample consensus: a paradigm for model fitting with applications to image analysis and automated cartography, 1981, Communications of the ACM, https://doi.org/10.1145/358669.358692.

ESTER, M., KRIEGEL, H.P., SANDER, J. AND XU, X. 1996: A Density-Based Algorithm for Discovering Clusters in Large Spatial Databases with Noise, 1996, Proceedings of the Second International Conference on Knowledge Discovery and Data Mining, 1996, https://doi.org/10.1145/358669.358692.

VAN DEN OORD, A., LI, Y., & VINYALS, O. 2019: Representation Learning with Contrastive Predictive Coding, 2019, https://arxiv.org/abs/1807.03748.

XIAO, A., HUANG, J., GUAN, D., CUI, K., LU, S., & SHAO, L. 2022: PolarMix: A General Data Augmentation Technique for LiDAR Point Clouds, 2022 Advances in Neural Information Processing Systems 35 (NIPS 2022).